\documentclass[letter, 10pt, conference]{IEEEtran}
\usepackage[letterpaper, left=0.75in, right=0.75in, bottom=0.75in, top=1in]{geometry}
\IEEEoverridecommandlockouts
\usepackage{cite}
\usepackage{amsmath,amssymb,amsfonts}

\usepackage[hidelinks]{hyperref} 
\def\sidenote{\footnote}
\usepackage[capitalise]{cleveref} 
\crefname{equation}{}{} 
\usepackage{flushend}

\usepackage{algorithm}
\usepackage{algpseudocode}
\newcommand{\lineref}[1]{line~\ref{#1}}

\usepackage{mathtools} 

\usepackage{graphicx}
\graphicspath{{images/}}
\usepackage[caption=false, font=footnotesize]{subfig} 

\usepackage{booktabs} 

\usepackage{textcomp}
\usepackage[dvipsnames]{xcolor}
\def\BibTeX{{\rm B\kern-.05em{\sc i\kern-.025em b}\kern-.08em
    T\kern-.1667em\lower.7ex\hbox{E}\kern-.125emX}}

\usepackage{todonotes}

\def\Cfree{\mathcal{C}_{\mathsf{free}}}
\def\V{\mathcal{V}}
\def\E{\mathcal{E}}
\def\Graph{\mathcal{G}}
\def\pic{\pi_c^{\text{sg}}}
\def\pid{\pi_d^{\text{sg}}}
\def\pidhat{\hat\pi{}_d^{\text{sg}}}
\def\Graph{\mathcal{G}}
\DeclareMathOperator*{\argmin}{arg\,min}
\newcommand{\norms}[1]{\lvert#1\rvert}
\newcommand{\normd}[1]{\lVert#1\rVert}
\newcommand{\length}[1]{\mathtt{length}\left(#1\right)}

\usepackage[nolist]{acronym}

\usepackage{siunitx}

\usepackage{zref-totpages}

\begin{document}

\begin{acronym}
    \acro{GSRM}{Gray-Scott Model Based Roadmap}
    \acro{ORM}{Optimized Roadmap Graph}
    \acro{PRM}{Probabilistic Roadmap}
    \acro{RRT}{Rapidly-exploring Random Tree}
    \acro{SPARS}{SPArse Roadmap Spanner}
    \acro{SPARS2}{SPArse Roadmap Spanner 2}
\end{acronym}

\title{GSRM: Building Roadmaps for Query-Efficient and Near-Optimal Path Planning Using a Reaction Diffusion System}

\author{
    Christian Henkel$^{1}$, Marc Toussaint$^{2}$, and Wolfgang H\"onig$^{2}$
    \thanks{$^{1}$Robert Bosch GmbH, Stuttgart, Germany \texttt{hec2le@bosch.com}}
    \thanks{$^{2}$TU Berlin, Germany 
    \texttt{\{toussaint, hoenig\}@tu-berlin.de}}
    \thanks{The research was partially funded by the Deutsche Forschungsgemeinschaft (DFG, German Research Foundation) - 448549715.}
    \thanks{© 2024 IEEE.  Personal use of this material is permitted.  Permission from IEEE must be obtained for all other uses, in any current or future media, including reprinting/republishing this material for advertising or promotional purposes, creating new collective works, for resale or redistribution to servers or lists, or reuse of any copyrighted component of this work in other works.}
}

\maketitle

\begin{abstract}
Mobile robots frequently navigate on roadmaps, i.e., graphs where edges represent safe motions, in applications such as healthcare, hospitality, and warehouse automation.
Often the environment is quasi-static, i.e., it is sufficient to construct a roadmap once and then use it for any future planning queries.
Roadmaps are typically used with graph search algorithm to find feasible paths for the robots.
Therefore, the roadmap should be well-connected, and graph searches should produce near-optimal solutions with short solution paths while simultaneously be computationally efficient to execute queries quickly.

We propose a new method to construct roadmaps based on the Gray-Scott reaction diffusion system and Delaunay triangulation.
Our approach, GSRM, produces roadmaps with evenly distributed vertices and edges that are well-connected even in environments with challenging narrow passages.
Empirically, we compare to classical roadmaps generated by 8-connected grids, probabilistic roadmaps (PRM, SPARS2), and optimized roadmap graphs (ORM).
Our results show that GSRM consistently produces superior roadmaps that are well-connected, have high query efficiency, and result in short solution paths.
\end{abstract}

\section{Introduction}

To use path finding algorithms such as A* for robot motion planning, a discretization of the environment is necessary.
A common approach is to discretize the environment as a grid.
However, such gridmaps generalize to non-rectangular environments ineffectively.
More generally, an environment can be discretized to a roadmap, which is a graph where vertices are locations that the robot can occupy, and the edges are possible motions between these locations.
A simple graph search in this roadmap will result in a feasible motion through the given environment.
The quality of the path is measured by its length and directly determined by the construction of the roadmap, assuming an optimal graph search algorithm.

The construction of a roadmap has multiple goals.
First, the roadmap should be well-connected, such that any two vertices are connected by a path.
Second, the roadmap should allow for efficient queries, i.e. it should have few vertices, because this improves the planning time when searching for a path.
Third, the roadmap should allow for short paths between any two vertices.
Finally, it is desirable to compute or update roadmaps efficiently.
Our work focuses mostly on the path length while maintaining connectedness and query-efficiency.
The roadmap it built once and then used for any path finding query.
Therefore, the query-efficiency and path length of the roadmap are more important properties than the computational efficiency of its construction.
While generally, roadmaps are often constructed in high-dimensional spaces, e.g., for manipulators, we focus this paper and the evaluation explicitly on the two-dimensional case that is common for mobile robots.

\begin{figure}
    \centering
    \includegraphics[width=.7\columnwidth, trim=36mm 13mm 32mm 14mm, clip=True]{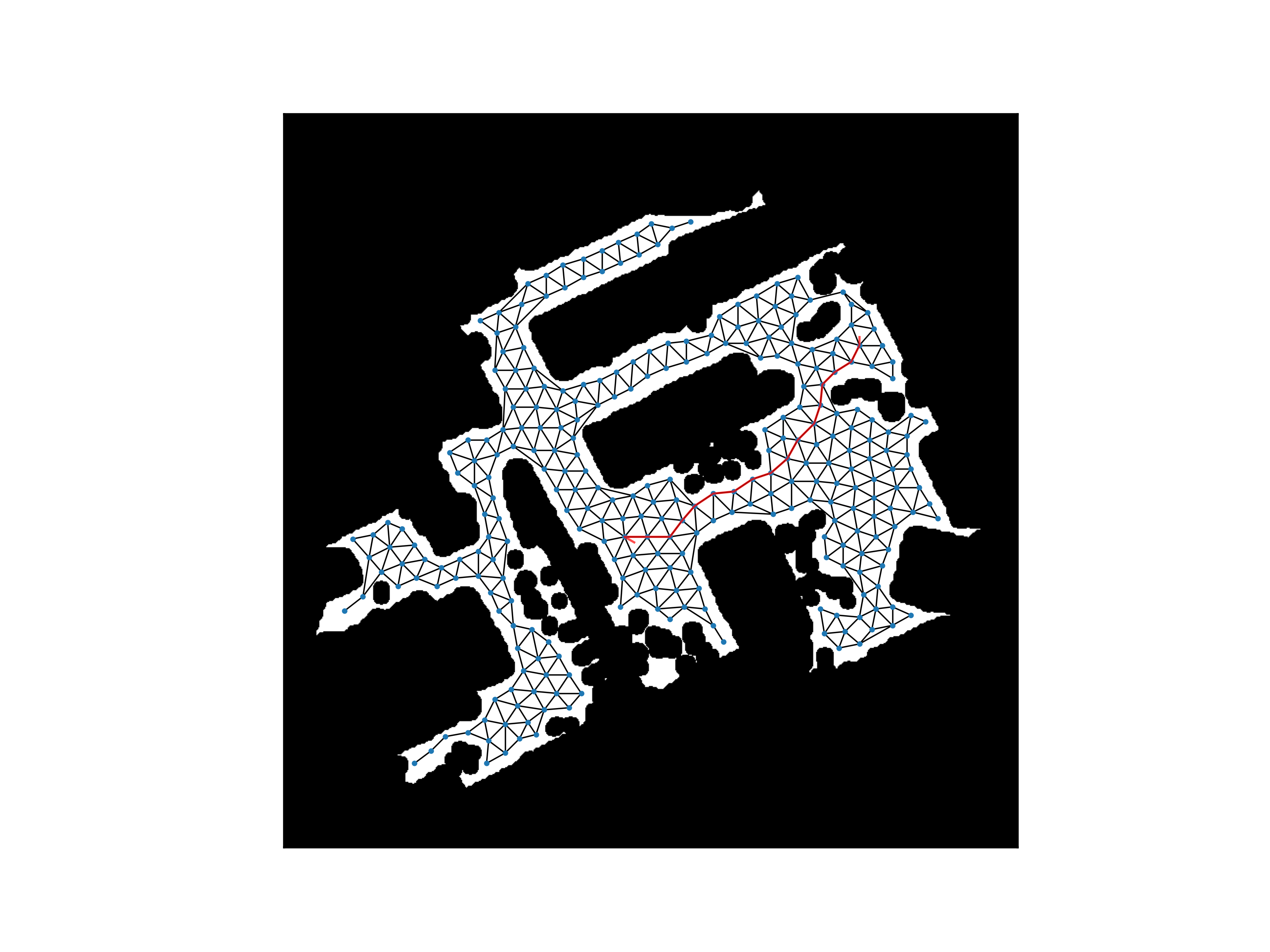}
    \caption{Example roadmap with $290$ nodes and $957$ edges on the map \emph{Slam} from \cref{fig:gsrm-eval-map-slam}.
    Blue dots are the vertices of the roadmap, black lines are the edges.
    The red line is an example path with a length of $0.479$.
    The roadmap was produced from the simulation of the Gray-Scott system shown in \cref{fig:gsrm-example-last}.}
    \label{fig:gsrm-example-graph}
\end{figure}

\begin{figure*}
    \centering
    \subfloat[At the beginning of optimization after random initialization.]{%
        \includegraphics[width=.24\textwidth]{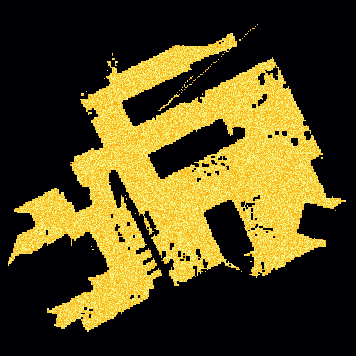}
    }
    \hfill
    \subfloat[After 10 simulation steps.]{%
        \includegraphics[width=.24\textwidth]{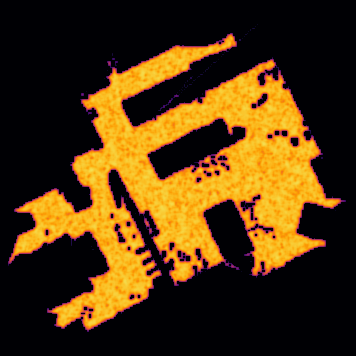}
    }
    \hfill
    \subfloat[After 500 simulation steps.]{%
        \includegraphics[width=.24\textwidth]{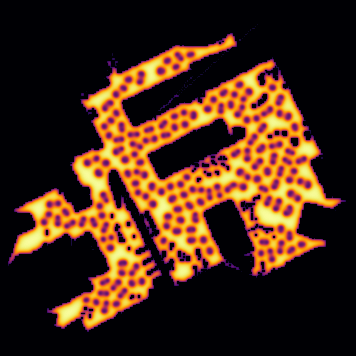}
    }
    \hfill
    \subfloat[After 10000 simulation steps.\label{fig:gsrm-example-last}]{%
        \includegraphics[width=.24\textwidth]{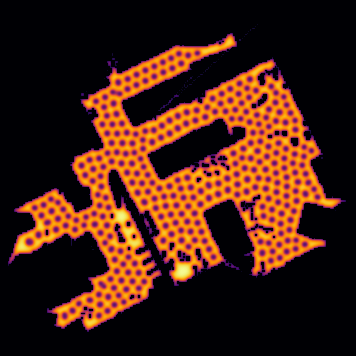}
    }
    \caption{Concentration of substance $U$ over time. The grid size is $l=300$. Black areas are obstacles, where the concentration is zero. The brighter the color, the higher the concentration of substance $U$. It is visible how the spotted pattern emerges.}
    \label{fig:gsrm-example}
\end{figure*}

A reaction diffusion system is a class of partial differential equations that model the diffusion of substances and their reaction with each other.
This model can produce spotted patterns \cite{chenStabilityDynamicsLocalized2011} that fill a given space, see \cref{fig:gsrm-example-last}.
It can be seen that these spots have a similar density throughout the accessible space.
This allows the generation of a roadmap graph by using the center points of these spots as vertices and connecting them locally by edges.
To find the edges, we use the Delaunay triangulation \cite{Delaunay1934a}.
The resulting roadmap is depicted in \cref{fig:gsrm-example-graph}.
We call this approach \acf{GSRM}.

\section{Related Work}
\label{sec:gsrm-related-work}
The basic idea of \acp{PRM} is to sample random points from the given configuration space and connect these with motions generated by a local planner \cite{Kavraki1996}.
These roadmaps are constructed once and can then be queried multiple times at runtime.
These queries can be performed with a classic graph search like A* \cite{hartFormalBasisHeuristic1968} to retrieve viable motions through the given configuration space.
A review over different \acp{PRM} is given in \cite{geraertsComparativeStudyProbabilistic2004}.

One version that is asymptotically optimal is called \ac{PRM}* \cite{Karaman2011a}.
Here, vertices are connected to more vertices compared to the traditional \ac{PRM}, to ensure asymptotical optimality. 
As a drawback this makes maps denser, which leads to more expensive path finding queries.
Since it is not delivering sparse roadmaps, we do not evaluate against it.

The alternative approach called \ac{SPARS} maintains near-optimality, but it additionally enforces the sparsity of the roadmap \cite{Dobson2013}.
It consists of one dense asymptotically optimal roadmap, similar to \ac{PRM}* and an additional spanner of that graph that maintains sparsity by only adding new points if necessary for path quality or connectivity.
\ac{SPARS}2 is an extension that improves the computational performance of the roadmap construction \cite{dobsonImprovingSparseRoadmap}.
We evaluate our approach against \ac{SPARS}2, because we consider it to be the state of the art in terms of sparsity and path length. 
We will show that our roadmap produces shorter paths with the same number of vertices.

Another work constructs a roadmap by focussing only on critical sections, e.g. doorways \cite{ichterLearnedCriticalProbabilistic2020}.
This can also lead to a sparse graph.
However, we want to construct a graph that spans the whole configuration space because we want to use it for the full path of the robot.

Roadmaps can also be built tailored to specific queries. 
\ac{RRT}, introduced by LaValle \cite{Lavalle1998rrt}, involves the incremental growth of a tree-like structure from an initial pose. 
This growth is achieved by systematically sampling potential new poses and verifying their connectivity within the environment.
In our research, we place particular emphasis on the multi-query approach to motion planning instead. 
Rather than constructing a new roadmap for each navigation task, we aim to leverage a roadmap that has been pre-computed and encodes information about the environment.

It is also common to construct roadmaps using Voronoi graphs \cite{chosetSensorBasedExplorationIncremental2000}.
For example, \cite{kallmannDynamicRobustLocal2014} proposes a method to build a roadmap by using so-called local clearance triangulations. 
This method can be used to generate paths with an arbitrary clearance.
There are also approaches to generate generalized Voronoi diagrams for sensor coverage \cite{tsardouliasConstructionMinimizedTopological2014}.
A method that uses Hamilton-Jacobi Skeletons for dynamic topological SLAM maps is \cite{modayilUsingTopologicalSkeleton2004}.
A similar approach can also be used to generate navigation roadmaps for robot exploration \cite{rezanejadRobustEnvironmentMapping2015}.
SphereMap \cite{musilSphereMapDynamicMultiLayer2022} builds a roadmap for UAV exploration in subterranean environments by sampling spherical segments.
A roadmap for exploration of mobile robots can be built by sampling disks and connecting them with a visibility graph \cite{noelDiskGraphProbabilisticRoadmap2022}.
We do not consider these methods in our evaluation, because they are optimized for area coverage of a sensor or for exploration, not for path planning.

Roadmaps can also be specifically constructed for multiple agents.
For example, by optimization \cite{Henkel2020} or hierarchical segmentation \cite{pratissoliHierarchicalTrafficManagement2023}.
In \cite{Debord2018}, \ac{SPARS} is used to plan trajectories for heterogeneous swarms of different aerial and ground robots.
\ac{ORM}, presented in \cite{Henkel2020} initializes the roadmap poses randomly and optimizes them by minimizing the length of randomly sampled paths.
We evaluate \ac{GSRM} against it in the evaluation section.
While generally, the multi-agent use case can also be solved by the roadmaps we propose, it is not the focus of this paper.

Reaction diffusion systems are a class of partial differential equations that model the diffusion of substances and their reaction with each other.
They can show complex patterns, also called \textit{Turing Patterns} \cite{turingChemicalBasisMorphogenesis1952}.
One of the most famous examples is the Gray-Scott system \cite{grayAutocatalyticReactionsIsothermal1984}.
When simulated in a grid, it produces spotted patterns \cite{chenStabilityDynamicsLocalized2011} as can be seen in \cref{fig:gsrm-example}.

Reaction diffusion systems have been used in robotics before, for example in robot navigation \cite{adamatzkyExperimentalReactionDiffusion2003, adamatzkyReactionDiffusionNavigationRobot2004,aidmanCoupledReactiondiffusionField2008, vazquez-oteroPathPlanningBased2012}, controller synthesis \cite{daleEvolutionReactiondiffusionControllers2010}, exploration \cite{vazquez-oteroReactionDiffusionBasedComputational2014}, and sensor coverage \cite{vazquez-oteroReactionDiffusionVoronoi2015}.

\section{Problem Formulation}
\label{sec:gsrm-problem-formulation}
Let $\Cfree \subset \mathbb{R}^2$ be the free space of the environment.
Let $\V \subset \Cfree$ be the vertices of the roadmap.
And let $\E \subseteq \V \times \V$ be the edges of the roadmap.
This forms the roadmap $\Graph = \langle \V, \E \rangle$.

We assume planning queries to be defined by a start location $s \in \Cfree$ and a goal location $g \in \Cfree$.
In $\Cfree$, we define the continuous path $\pic: \mathbb{R} \to \Cfree$ from $s$ to $g$.
We use the vertices closest to $s$ and $g$ as start vertex $v_s$ and goal vertex $v_g$, respectively, i.e.
\begin{equation}
    \begin{aligned}
        v_s & = \argmin_{v \in \V} \normd{v - s}  \\
        v_g & = \argmin_{v \in \V} \normd{v - g},
    \end{aligned}
\end{equation}
where $\normd{\cdot}$ is the Euclidean distance.

The discrete path $\pid: \{0, 1, \dots, n\} \to \V$ is defined by $\pid(0) = v_s$ and $\pid(n) = v_g$, where
\begin{equation}
    \forall i \in \{0, 1, \dots, n-1\}, \langle\pid(i), \pid(i+1)\rangle \in \E.
\end{equation}
Its length is defined as
\begin{equation}
    \length{\pid} := \sum_{i=0}^{n-1} \normd{\pid(i+1) - \pid(i)}.
\end{equation}
Let $\Pi_d^{\text{sg}}$ be the set of all discrete paths from $v_s$ to $v_g$. Then the shortest path between $v_s$ and $v_g$ is defined as
\begin{equation}
    \pidhat = \argmin_{\pid \in \Pi_d^{\text{sg}}} \length{\pid}.
\end{equation}
We define the length of the path $\pic$ in $\Cfree$ as the geometric length of the shortest discrete path which includes the distance from $s$ to the first vertex and the distance from the last vertex to $g$:
\begin{equation}
    \length{\pic} := \length{\pidhat} + \normd{\pidhat(0) - s} + \normd{\pidhat(n) - g}.
    \label{eq:gsrm-path-length-free}
\end{equation}

The goal of the roadmap construction is to find a roadmap $\Graph$ that minimizes the expected length of the paths while limiting the number of vertices in the graph to a maximum number $n_{\text{max}}$:

\begin{equation}
    \argmin_{\Graph} \int\limits_{\mathclap{(s,g) \in \Cfree\times \Cfree}} P(s, g) \length{\pic} \text{~s.t.~} \norms{\V} < n_{\text{max}}.
    \label{eq:gsrm-argmin}
\end{equation} 

In this paper, we assume $P(s, g)$ to be a uniform distribution over $\Cfree \times \Cfree$.

\section{Gray-Scott System}
\label{sec:gsrm-gray-scott-system}

To generate the roadmap we first need to define vertex positions in free space.
For this, we use the Gray-Scott system \cite{grayAutocatalyticReactionsIsothermal1984}.
The Gray-Scott system is a simplified model of a chemical reaction-diffusion system.
It models two chemical substances that react with each other and move by diffusion.
The substances are generally called $U$ and $V$ and their chemical reaction is defined as
\begin{equation}
    U + 2V \to 3V,
\end{equation}
where $x \to y$ means that reactants $x$ yield products $y$.

Additionally, the system models a constant addition of substance $U$ according to the feed rate $A$ as well as a constant removal of $V$ according to the kill rate $B$.
The diffusion of the substances is quantified by the diffusion parameters $D_U$ and $D_V$ respectively.

For a certain configuration of the aforementioned variables $A$, $B$, $D_U$, and $D_V$, it produces stable spotted patterns that fill a given two-dimensional space \cite{pearsonComplexPatternsSimple1993}.

\subsection{Simulation}
We simulate this system in a grid of $l \times k$ cells, where $u \in \mathbb{R}^{l \times k}$ and $v \in \mathbb{R}^{l \times k}$ are the concentration of substance $U$ and $V$, respectively.
The \textit{\acf{GSRM} Construction} in \cref{alg:gsrm_construction} describes the simulation of the Gray-Scott system.
We will elaborate the algorithm in the following.

\begin{algorithm}
    \begin{algorithmic}[1]
        \State \textbf{Input:} $A$, $B$, $D_U$, $D_V$, $l$, $k$, $\Cfree$, $u_l$, $u_h$, $v_l$, $v_h$
        \State \textbf{Output:} $\V$, $\E$
        \State $u \gets \mathtt{random\_uniform}(l, k, u_l, u_h)$ \label{alg:ln:gsrm-init-u}
        \State $v \gets \mathtt{random\_uniform}(l, k, v_l, v_h)$ \label{alg:ln:gsrm-init-v}
        \For {$[0, N]$}
        \State $\forall (x, y) \notin \Cfree: u(x, y) \gets 0$ \label{alg:ln:gsrm-obstacle}
        \State $\forall (x, y) \notin \Cfree: v(x, y) \gets 0$
        \State $\frac{\delta u}{\delta t} \gets D_U \Delta u - u v^2 + A(1-u)$ \label{alg:ln:gsrm-delta-u}
        \State $\frac{\delta v}{\delta t} \gets D_V \Delta v + u v^2 - (A + B)v$ \label{alg:ln:gsrm-delta-v}
        \State $u \gets u + \frac{\delta u}{\delta t}$ \label{alg:ln:gsrm-update-u}
        \State $v \gets v + \frac{\delta v}{\delta t}$ \label{alg:ln:gsrm-update-v}
        \EndFor
        \State $v_{\text{threshold}} \gets \left\{\def\arraystretch{1.2}
            \begin{tabular}{@{}l@{\quad}l@{}}
                $1$ & if $v(x, y) > \frac{\max v}{2}$   \\
                $0$ & if $v(x, y) \le \frac{\max v}{2}$
            \end{tabular}\right.$ \label{alg:ln:gsrm-threshold}
        \State $\mathtt{contours} \gets \mathtt{find\_contours}(v_{\text{threshold}})$ \label{alg:ln:gsrm-contours}
        \State $\V \gets \varnothing$  
        \For {$\mathtt{contour} \in \mathtt{contours}$}
        \State $\mathtt{center} \gets \mathtt{average}(\mathtt{contour})$ \label{alg:ln:gsrm-center}
        \State $\V \gets \V \cup \{\mathtt{center}\}$
        \EndFor
        \State $\V_{\text{dummy}} \gets \mathtt{make\_vertices\_in\_obstacles}(\Cfree)$  \label{alg:ln:gsrm-dummy}
        \State $\E \gets \varnothing$  
        \State $\mathtt{triangles} \gets \mathtt{delaunay}(\V \cup \V_{\text{dummy}})$ \label{alg:ln:gsrm-delaunay}
        \For {$(\mathtt{a}, \mathtt{b}, \mathtt{c}) \in \mathtt{triangles}$} \label{alg:ln:gsrm-edge-start}
        \For {$\mathtt{e} \in \{(\mathtt{a}, \mathtt{b}), (\mathtt{b}, \mathtt{c}), (\mathtt{c}, \mathtt{a})\}$}
        \If {$\mathtt{free}(\mathtt{e}, \Cfree)$}
        \State $\E \gets \E \cup \{\mathtt{e}\}$
        \EndIf
        \EndFor 
        \EndFor \label{alg:ln:gsrm-edge-end}
        \State \textbf{return} $\V$, $\E$
    \end{algorithmic}
    \caption{\acf{GSRM} Construction}
    \label{alg:gsrm_construction}
\end{algorithm}

\begin{figure*}[ht!]
    \centering
    \subfloat[Map Plain.\label{fig:gsrm-eval-map-plain}]{%
        \includegraphics[width=.24\textwidth]{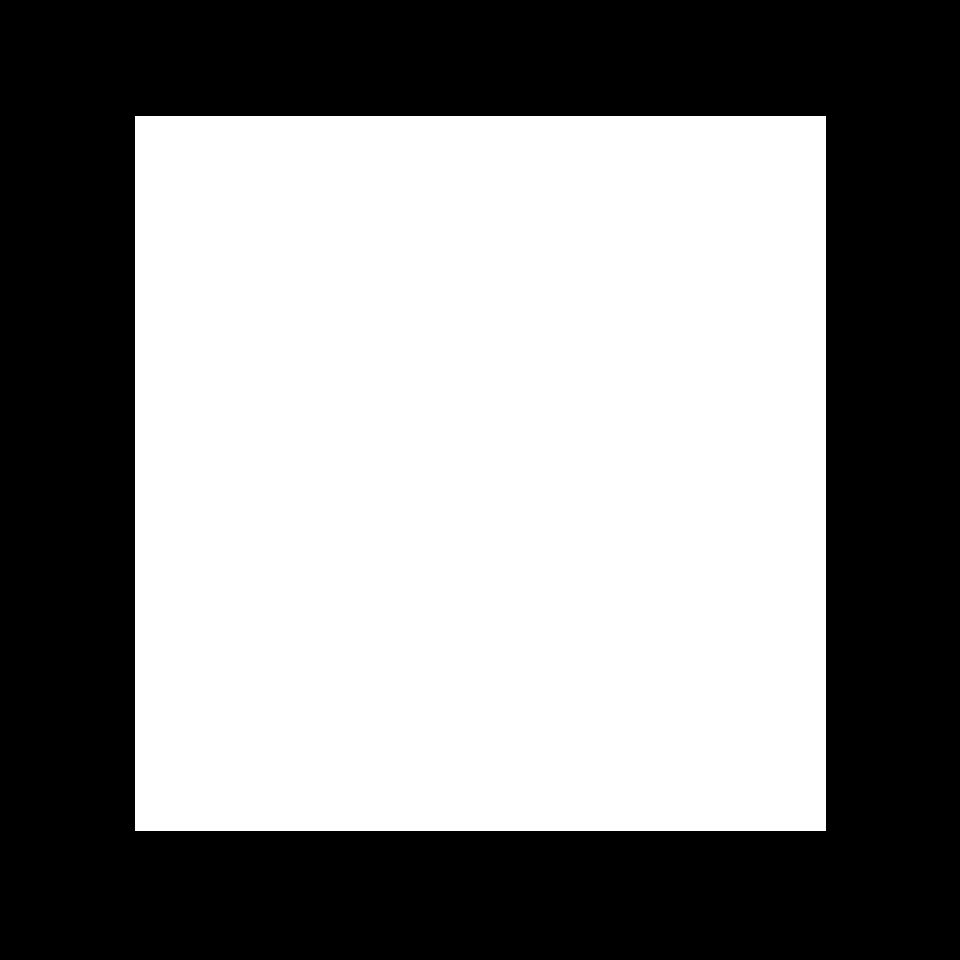}
    }
    ~
    \subfloat[Map Den.\label{fig:gsrm-eval-map-brc}]{%
        \includegraphics[width=.24\textwidth]{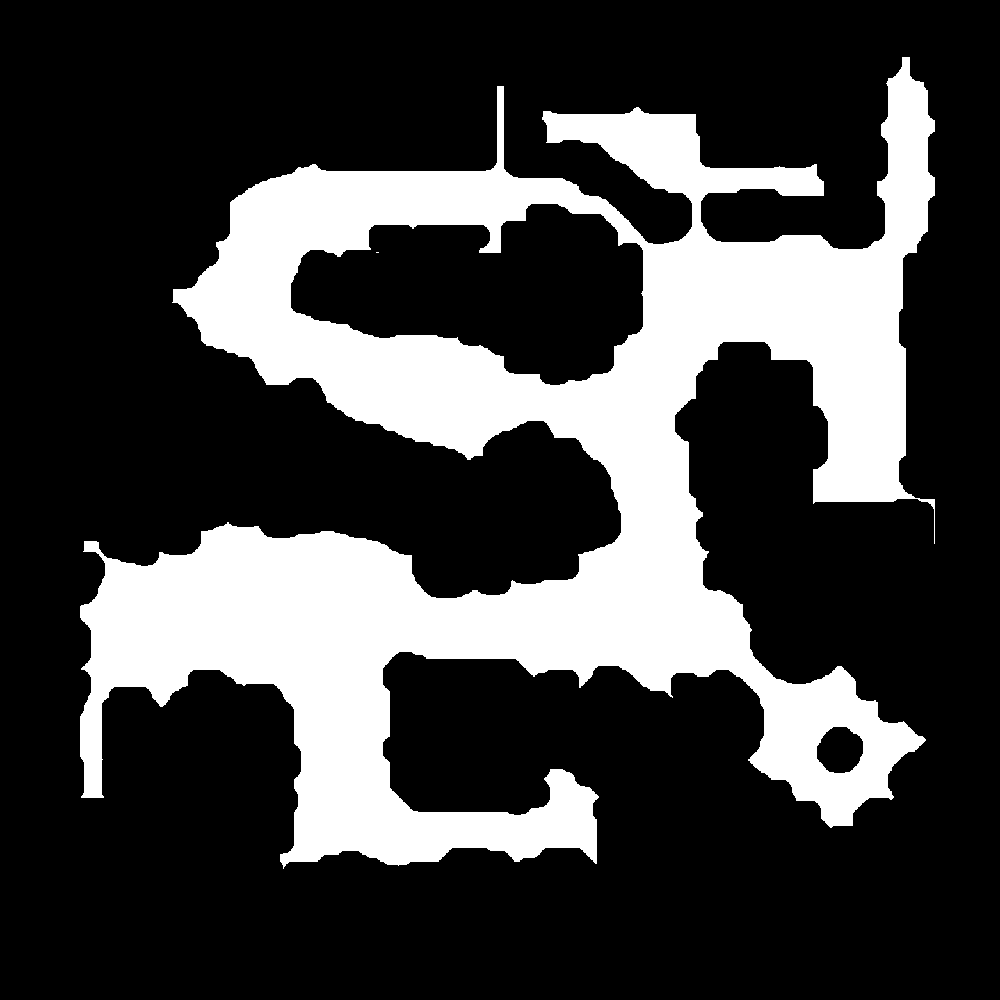}
    }
    ~
    \subfloat[Map Rooms.\label{fig:gsrm-eval-map-rooms}]{%
        \includegraphics[width=.24\textwidth]{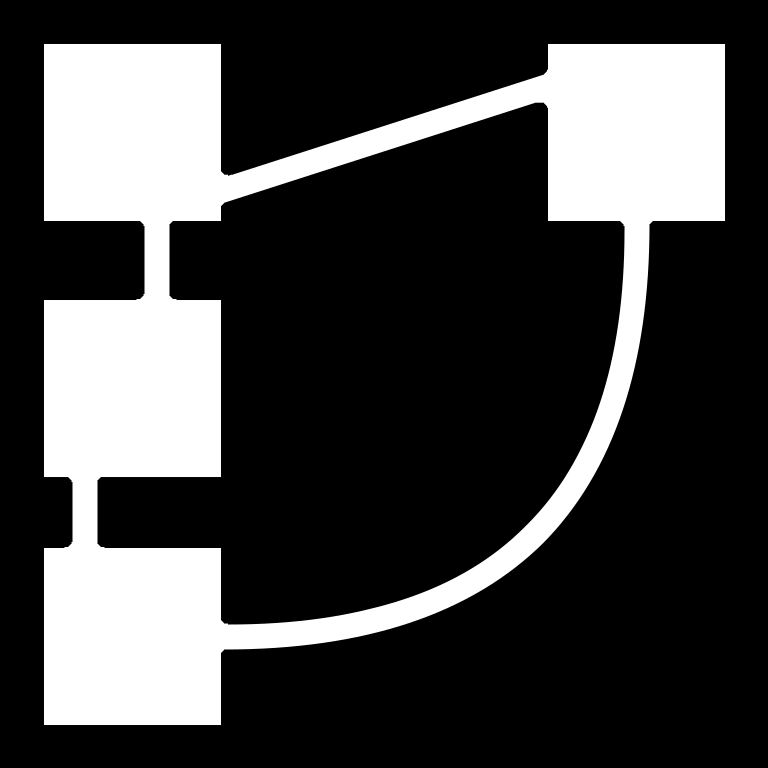}
    }
    ~
    \subfloat[Map Slam.\label{fig:gsrm-eval-map-slam}]{%
        \includegraphics[width=.24\textwidth]{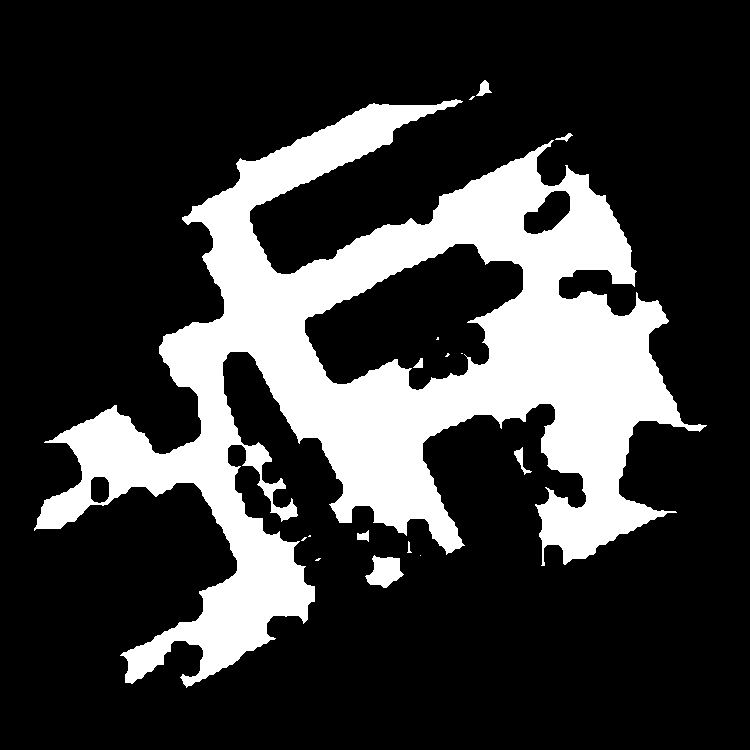}
    }
    \caption{Maps used in evaluation. Gray and black areas are obstacles. White areas are free space.}
    \label{fig:gsrm-eval-maps}
\end{figure*}

Let $u_l$ and $u_h$ be the lower and upper bounds of the concentration of substance $U$ and $v_l$ and $v_h$ be the lower and upper bounds of the concentration of substance $V$.
We initialize $u$ with values between $u_l$ and $u_h$ in \lineref{alg:ln:gsrm-init-u} and $v$ with values between $v_l$ and $v_h$ in \lineref{alg:ln:gsrm-init-v}.
The function $\mathtt{random\_uniform}\left(l, k, x, y\right)$ returns an $l \times k$ matrix filled with values, each independently sampled from the uniform distribution of values between $x$ and $y$.

At the borders of the grid and in obstacles, we set both concentrations to zero in \lineref{alg:ln:gsrm-obstacle}.
This stops the pattern from spreading into these regions. 
In \lineref{alg:ln:gsrm-delta-u} and \lineref{alg:ln:gsrm-delta-v}, we calculate the update for the concentrations of the substances according to the following differential equations:
\begin{equation}
    \begin{aligned}
        \frac{\delta u}{\delta t} & = D_U \Delta u - u v^2 + A(1-u)    \\
        \frac{\delta v}{\delta t} & = D_V \Delta v + u v^2 - (A + B)v,
    \end{aligned}
\end{equation}
where $\Delta$ is the Laplace operator and all other operations are performed element-wise.

The update is performed in a loop, $N$ times.
See \cref{fig:gsrm-example} for an example of the concentration of substance $U$ at different states of the simulation.

\subsection{Identifying Vertices}
\label{sec:gsrm-identifying-vertices}

Based on the result of the simulation of the Gray-Scott system, we have to identify the separate spots in \cref{fig:gsrm-example-last} and use them as vertices $\V$ of the roadmap to produce a result as shown in \cref{fig:gsrm-example-graph}.
First, we transform the concentration of substance $V$ into a binary image by thresholding it at half of the maximum concentration in \lineref{alg:ln:gsrm-threshold} of \cref{alg:gsrm_construction}.
Then we apply the border following algorithm \cite{suzukiTopologicalStructuralAnalysis1985} to separate the spots in \lineref{alg:ln:gsrm-contours}. 
The function $\mathtt{find\_contours}\left(x\right)$ is called on the binary image $x$ and returns a list of unique contours, each represented as a list of points.
The lists include all the outermost points per contour, i.e. those that have the value $1$ and at least one neighbor with the value $0$.

Based on the contours, we estimate the center of mass of each spot.
We take the average over all points of the contour in \lineref{alg:ln:gsrm-center}.
Here, the method $\mathtt{average}(\mathtt{contour})$ takes the points that are part of the contour of the given spot and return their centroid.
The result is a set of points $\V$ that are the vertices of the roadmap.

To avoid artifacts in the following step, we need to additionally create dummy vertices in \lineref{alg:ln:gsrm-dummy}.
These are vertices that are located in the non-free areas of the environment.
If they would not exist, the following Delaunay triangulation would produce a lot of small, acute triangle in the areas close to obstacles.
However, these vertices will not be considered in the final roadmap.

\subsection{Identifying Edges}
\label{sec:gsrm-identifying-edges}

The next step is to identify the edges $\E$ of the roadmap.
We use the Delaunay triangulation \cite{Delaunay1934a} for this purpose.
The result is a set of triangles calculated in \lineref{alg:ln:gsrm-delaunay} of \cref{alg:gsrm_construction}.
In lines \ref{alg:ln:gsrm-edge-start} to \ref{alg:ln:gsrm-edge-end} we add an edge between each pair of vertices that are connected by a triangle side iff the edge $(a, b)$ is free, meaning that the straight line between $a$ and $b$ is a subset of $\Cfree$:
\begin{equation}
    \mathtt{free}\left((a, b), \Cfree\right) = (tb + (1-t)a) \in \Cfree, \forall t \in [0, 1].
\end{equation}

\section{Evaluation}
\label{sec:gsrm-evaluation}

The \acf{GSRM} algorithm is implemented in C++ using methods from the \emph{OpenCV} library\sidenote{\url{https://opencv.org/}}.
The source code is publicly available\sidenote{\url{https://ct2034.github.io/miriam/iros2024/}}.

For the evaluation we rely on four maps with different properties, that are all located in the unit square $[0, 1]^2$:

\begin{itemize}
    \item Map \textbf{Plain} in \cref{fig:gsrm-eval-map-plain} is an open square with no obstacles.
          This map demonstrates how the roadmaps allow the agents to move in an open space.
    \item Map \textbf{Den} in \cref{fig:gsrm-eval-map-brc} is from a path-finding benchmark \cite{Stern2019}.
          Its special property is that it has comparatively organic shapes and we use it to evaluate the performance of the roadmaps in a more complex environment.
    \item Map \textbf{Rooms} in \cref{fig:gsrm-eval-map-rooms} is a map with 4 rooms connected by narrow corridors.
          One of the corridors is curved and we want to check how and if the roadmaps can connect through it.
    \item Map \textbf{Slam} in \cref{fig:gsrm-eval-map-slam} is a map that was achieved by running \emph{SLAM} on a real robot in a household environment.
          It was included to evaluate the performance of the roadmaps in a real-world environment.
\end{itemize}

\begin{figure*}
    \subfloat[Example of a roadmap with $290$ vertices and $963$ edges generated by the \ac{GSRM} algorithm. The path has a length of $1.134$, which is the shortest path of all evaluated roadmaps.
    \label{fig:gsrm-roadmap}]{
        \includegraphics[width=.31\textwidth, trim=36mm 13mm 32mm 14mm, clip=true]{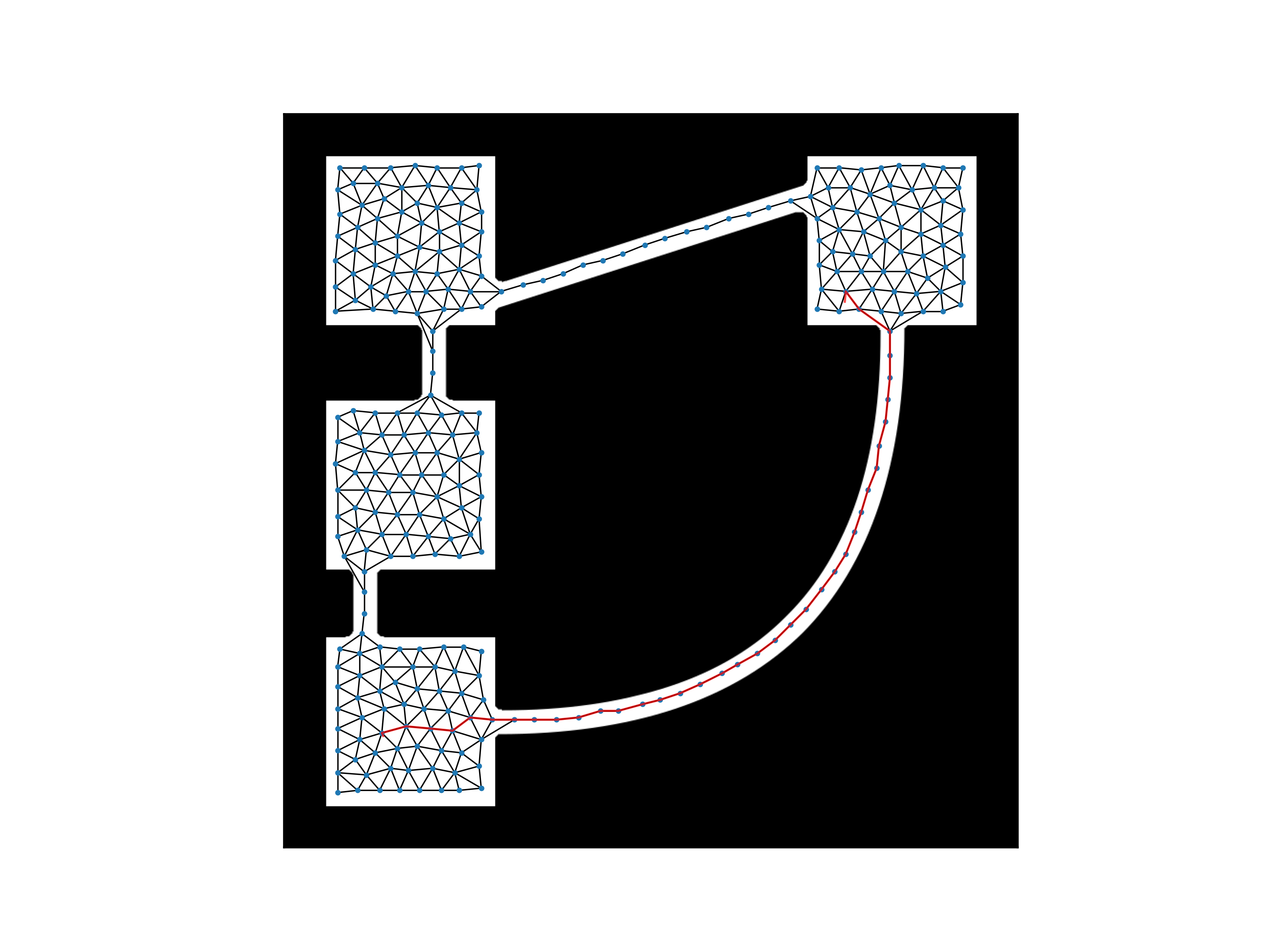}
    }
    ~
    \subfloat[Example of a roadmap with $272$ vertices and $520$ edges generated by \ac{SPARS}2. The path has a length of $1.401$.
    \label{fig:gsrm-other-spars}]{
        \includegraphics[width=.31\textwidth, trim=36mm 13mm 32mm 14mm, clip=true]{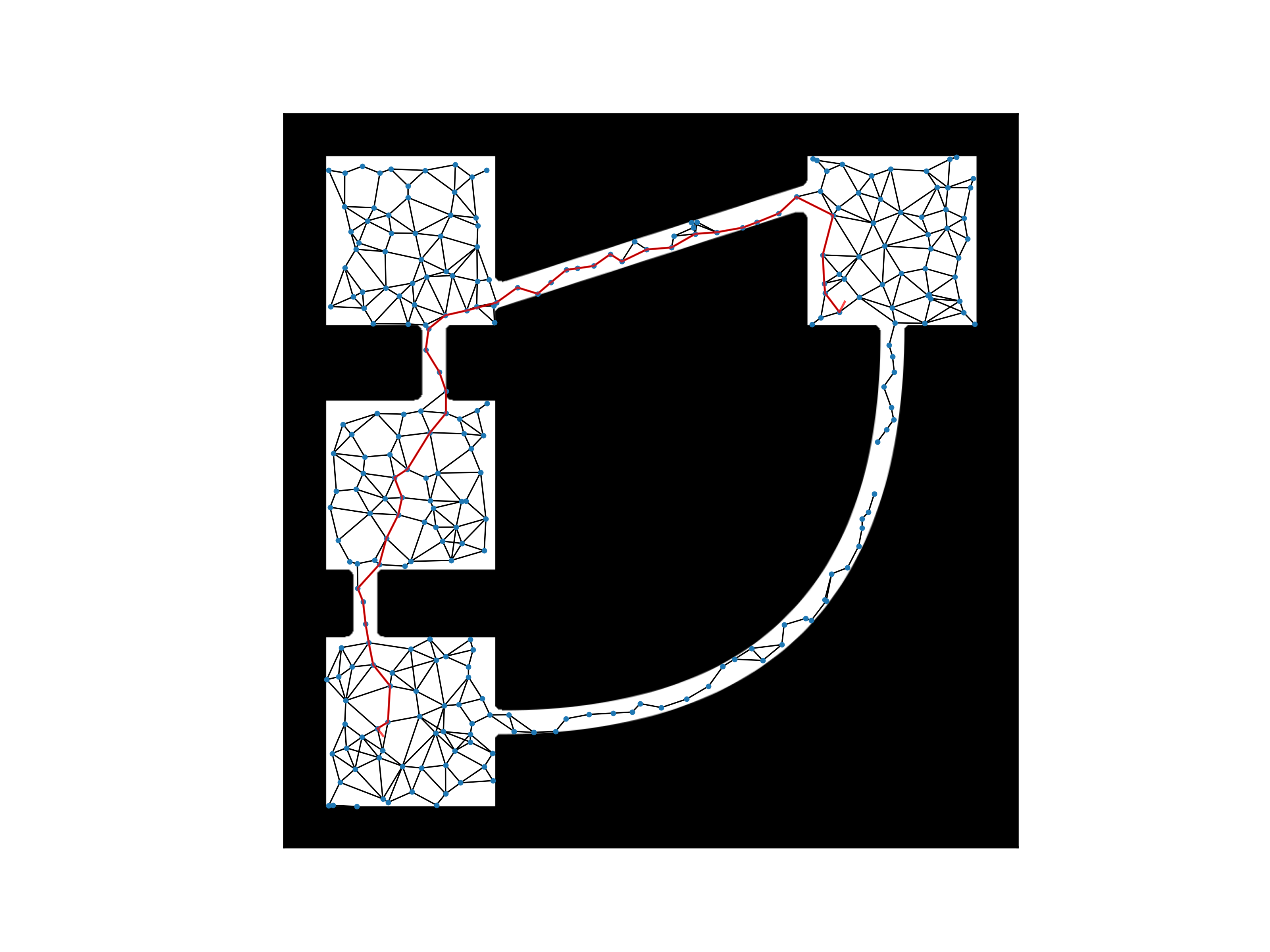}
    }
    ~
    \subfloat[Example of a roadmap with $270$ vertices and $932$ edges generated by ORM. The path has a length of $1.381$.
    \label{fig:gsrm-other-orm}]{
        \includegraphics[width=.31\textwidth, trim=36mm 13mm 32mm 14mm, clip=true]{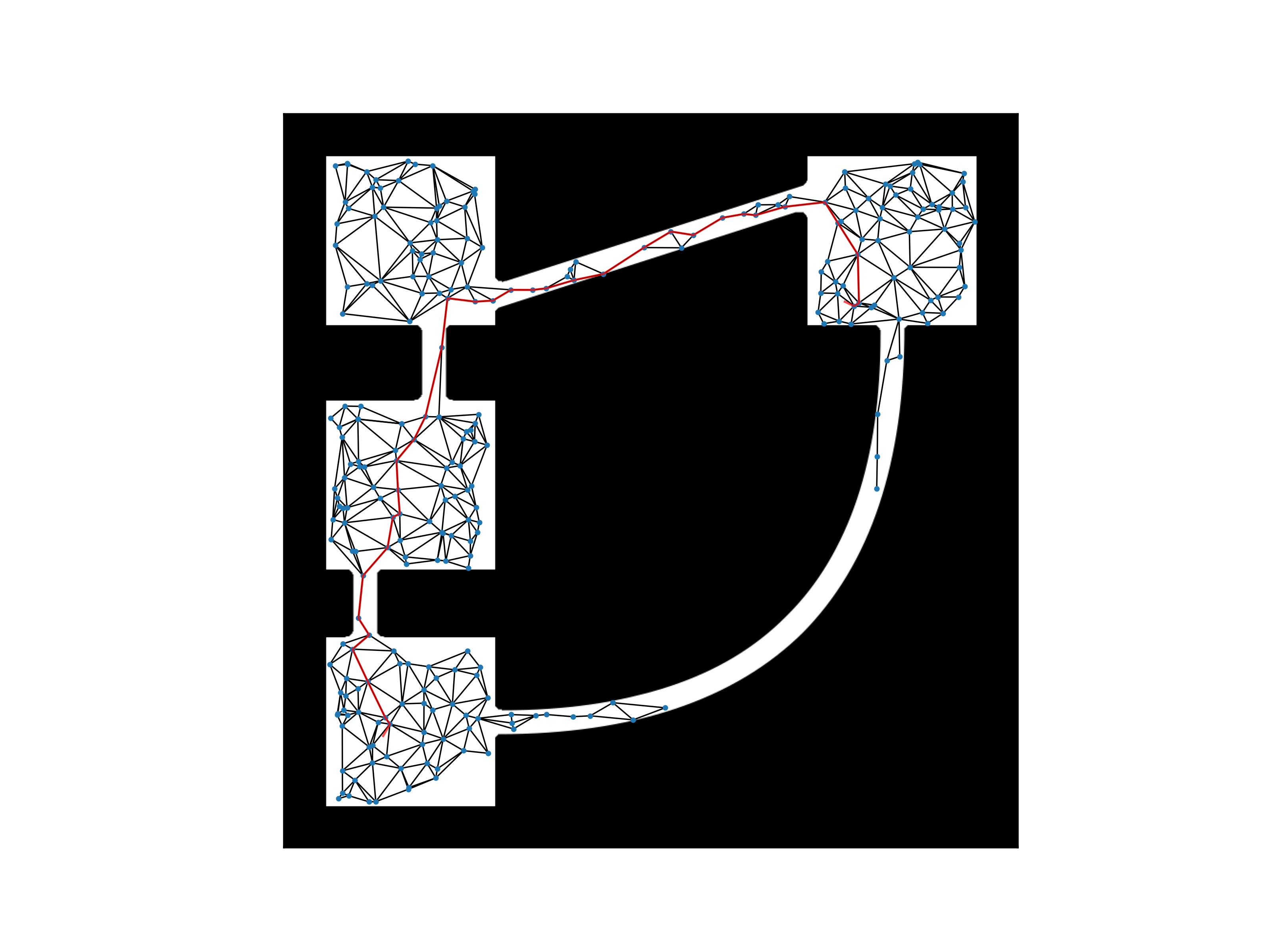}
    }
    \\
    \subfloat[Example of a roadmap with $282$ vertices and $1207$ edges generated by \ac{PRM}. The path has a length of $0.448$.
    \label{fig:gsrm-other-prm}]{
        \includegraphics[width=.31\textwidth, trim=36mm 13mm 32mm 14mm, clip=true]{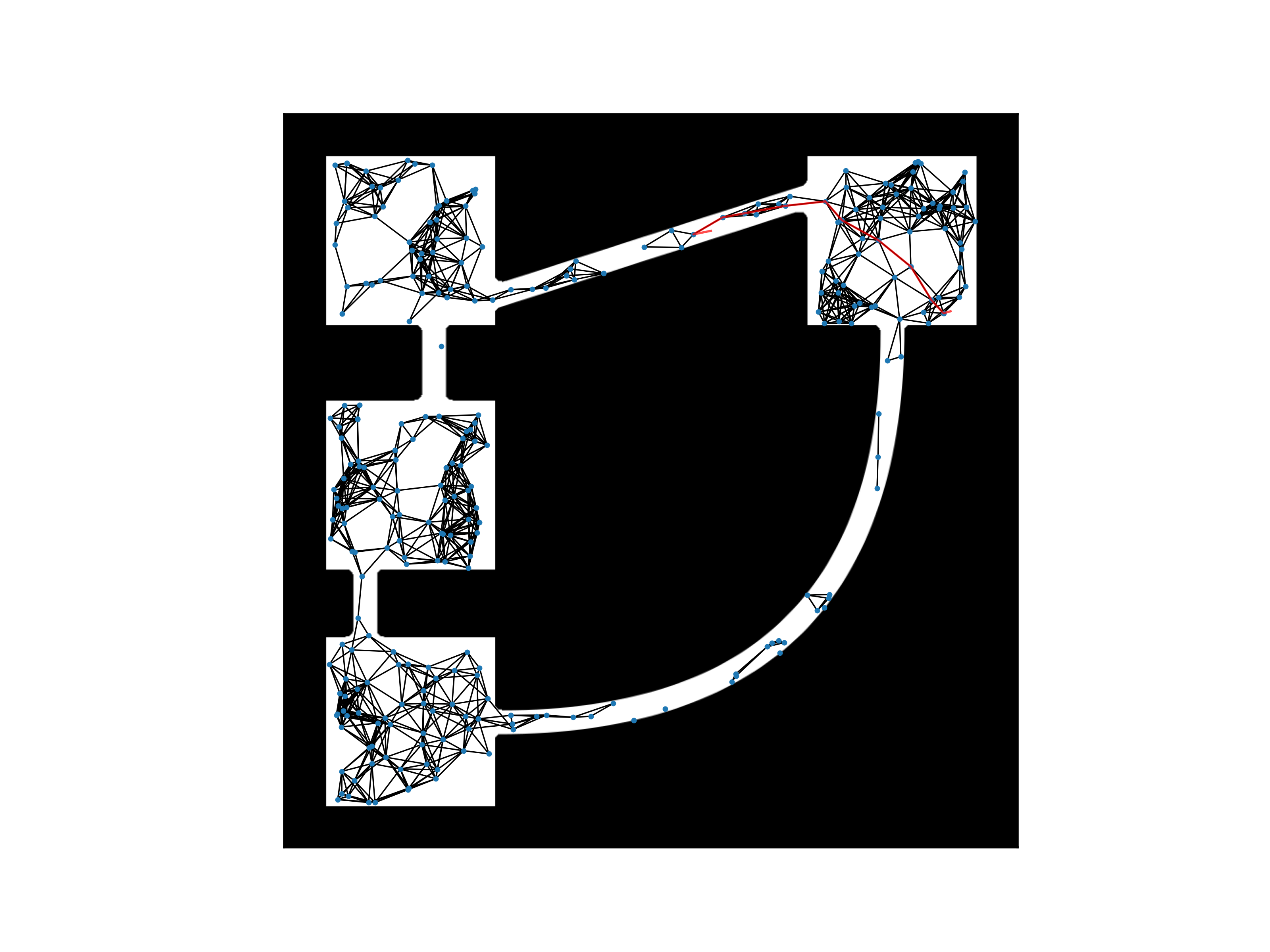}
    }
    ~
    \subfloat[Example of a gridmap with $298$  vertices and $877$ edges. The path has a length of $0.413$.
    \label{fig:gsrm-other-gridmap}]{
        \includegraphics[width=.31\textwidth, trim=36mm 13mm 32mm 14mm, clip=true]{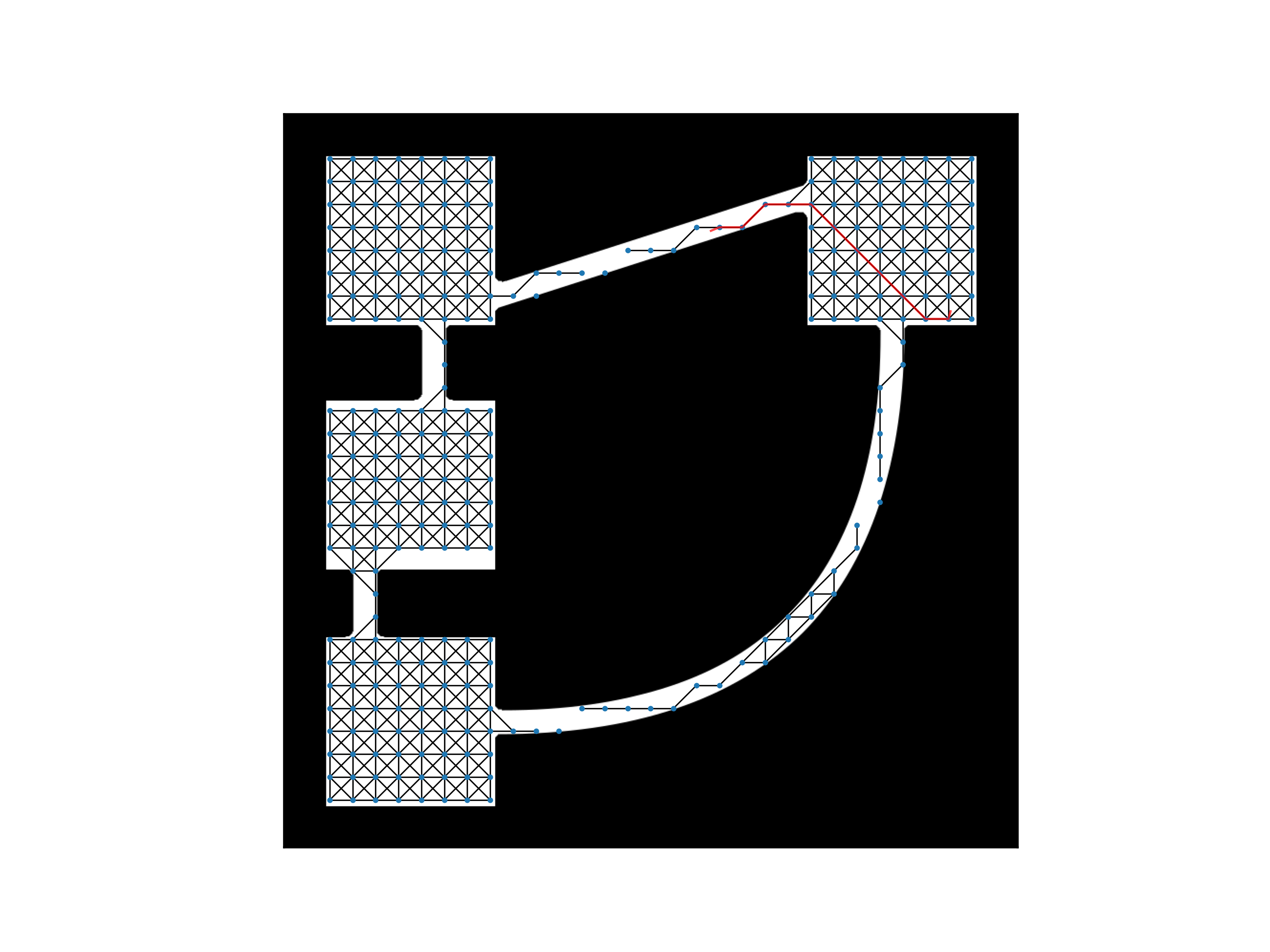}
    }
    
    \caption{Examples of roadmaps generated by different algorithms in map \emph{Rooms}. The red line is an example path. The paths for \ac{GSRM}, SPARS2, and ORM share the same start and goal points, while the paths for \ac{PRM} and Gridmap also share the same start and goal points. The last two examples use a different path query, because the query from the first examples are not successfully computable with them.}
    \label{fig:gsrm-eval-algorithms}
\end{figure*}

\begin{table*}
\caption{Parameters for the \ac{GSRM} algorithm.}
    \begin{center}
        \begin{tabular}{l|lllllllll}
            \toprule
            Parameter & $N$    & $D_U$ & $D_V$ & $A$   & $B$   & $u_l$ & $u_h$ & $v_l$ & $v_h$ \\
            \midrule
            Value     & 10 000 & 0.14  & 0.06  & 0.035 & 0.065 & 0.8   & 1.0   & 0.0   & 0.2   \\
            \bottomrule
        \end{tabular}
    \end{center}
    \label{tab:gsrm-eval-parameters}
\end{table*}

We compare the \ac{GSRM} algorithm with the following algorithms:
\begin{itemize}
    \item \textbf{\acf{SPARS2}} \cite{dobsonImprovingSparseRoadmap}.
          \cref{fig:gsrm-other-spars} shows an example of a roadmap generated by \ac{SPARS}2.
          Note that we did not find hyperparameters for \ac{SPARS}2 to generate roadmaps with as small number of vertices as we achieved with the other algorithms.
    \item \textbf{\acf{ORM}}, a variant of \cite{Henkel2020} where all edges are undirected.
          \cref{fig:gsrm-other-orm} shows an example of a roadmap generated by \ac{ORM}.
    \item \textbf{\acf{PRM}} \cite{Kavraki1996}.
          \cref{fig:gsrm-other-prm} shows an example of a roadmap generated by \ac{PRM}.
          We used an implementation where vertices are connected if they are within a distance $\delta$ of each other.
          We set $\delta$ such that the number of edges is the same as for \ac{GSRM}.
    \item \textbf{Gridmap}, an 8-connected gridmap that has all vertices and edges removed that are in obstacles.
          \cref{fig:gsrm-other-gridmap} shows an example of a gridmap.
\end{itemize}

For all algorithms, we have chosen sets of parameters such that the number of vertices is similar across the algorithms.
In the \ac{GSRM} algorithm, we can control the number of vertices by changing the resolution $l = k$.
For the other parameters, see \cref{tab:gsrm-eval-parameters}.
These values are necessary to produce a robust spotted pattern in the simulation.

For the evaluation, we approximate the distribution in \cref{eq:gsrm-argmin} numerically by a $100$ discrete pairs of points per configuration.

\subsection{Path length}
\label{sec:gsrm-evaluation-path-length}
We evaluate the quality of the roadmap by measuring the path length of the shortest path between two random points in $\Cfree$.
For all roadmap configurations $10$ roadmaps were built.
All roadmaps are evaluated with the same $100$ pairs of points.
The path length is evaluated according to \cref{eq:gsrm-path-length-free}.
To find the shortest path in the roadmap graph, we use the \emph{A*} algorithm \cite{hartFormalBasisHeuristic1968} from the \emph{Boost} library\sidenote{https:\/\/www.boost.org\/} with the Euclidean distance as heuristic.

\begin{figure}
    \centering
    \includegraphics[width=\columnwidth]{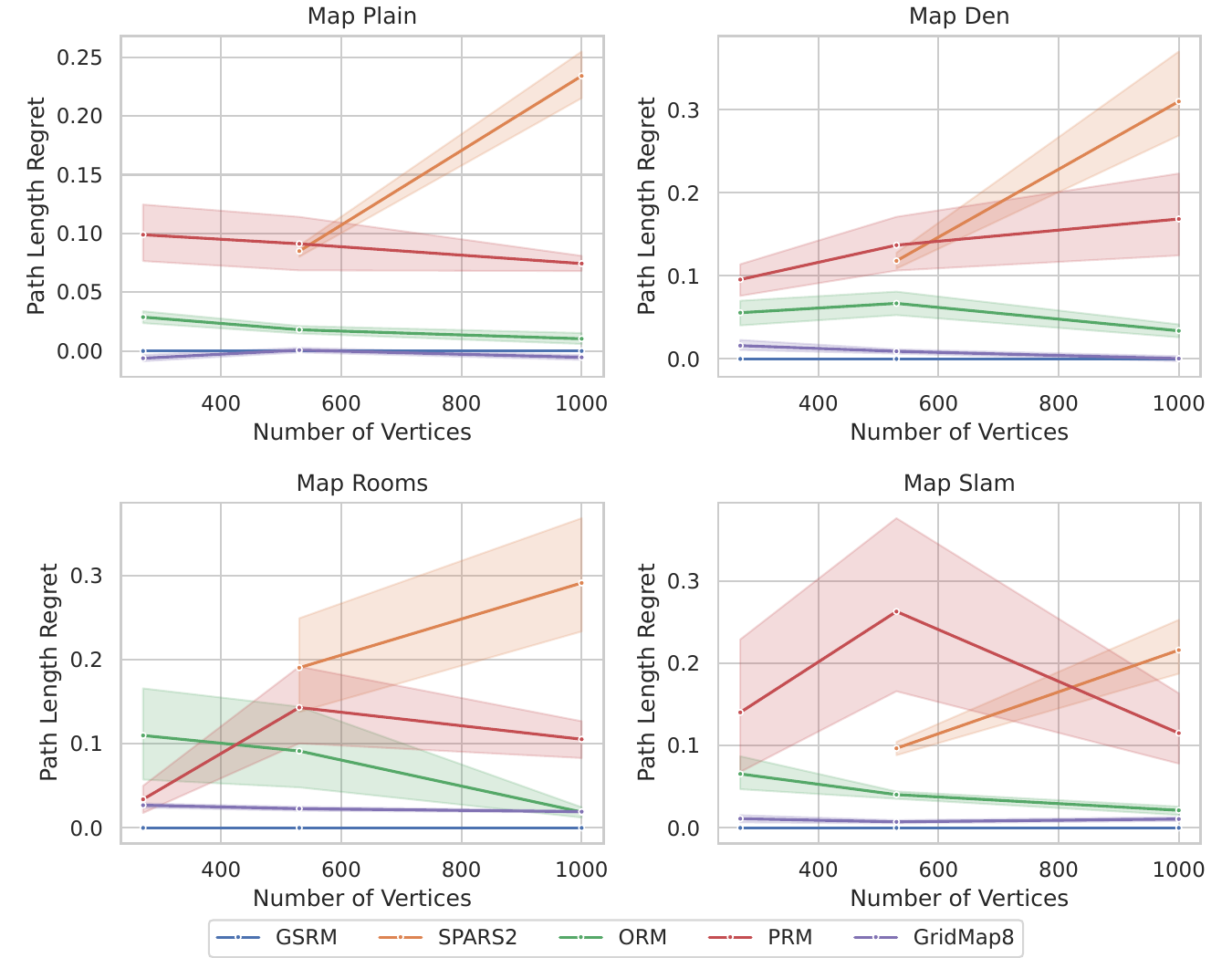}
    \caption{Average path length regret of the shortest path between two random points in $\Cfree$ for different roadmap types. The shaded areas show the standard deviation across $100$ trials. \ac{GSRM} provides the shortest paths in maps other than \emph{Plain}.}
    \label{fig:gsrm-path-length}
\end{figure}



\cref{fig:gsrm-path-length} shows the average path length regret of the shortest path between two random points in $\Cfree$ for different roadmap types on different maps.
The regret per datapoint is calculated by $$\mathtt{regret} = (\mathtt{pl}_{\text{other}} - \mathtt{pl}_{\text{GSRM}}) / \mathtt{pl}_{\text{other}},$$ where $\mathtt{pl}_{\text{other}}$ refers to the path length produced on the respective other roadmap, while $\mathtt{pl}_{\text{GSRM}}$ refers to the one on our roadmap.

It shows that the \ac{GSRM} algorithm generates roadmaps with a lower path length than the other algorithms in all maps except \emph{Plain}.
In the map \emph{Plain}, the 8-connected gridmap produces shorter paths, because it is denser in terms of edges per vertex.
Maps that are more structured, such as \emph{Den} and \emph{Rooms}, show a larger difference in path length between the \ac{GSRM} algorithm and the other algorithms.
This is because the \ac{GSRM} algorithm can generate roadmaps that adapt to the structure of the environment, while especially the 8-connected gridmap is not able to do so.
Generally, for most roadmaps, the path length regret decreases with the number of vertices.
The benefit of the \ac{GSRM} algorithm is that it can generate roadmaps with a low path length even with a low number of vertices.

For the roadmaps PRM and SPARS2 it is visible that the regret increases with a higher number of vertices.
This is counterintuitive because it is generally expected that higher numbers of vertices lead to shorter paths. 
But it can be explained by taking the success rates into account: If less path queries are successful, those that are still successful may result in longer paths.

\subsection{Success Rate}
\label{sec:gsrm-evaluation-success-rate}
We evaluate the success rate of the roadmap by measuring the fraction of successful path planning queries.
As discussed before, paths are planned between two random points in $\Cfree$.
A path then includes the section from the start point to the first vertex and from the last vertex to the goal point.
We consider a path unsuccessful if there is no path connecting the two vertices in the roadmap or if there is no free path on the line between the start point and the first vertex or the last vertex and the goal point.
This was also evaluated for the $100$ same pairs of start and goal configurations per roadmap on $10$ roadmaps per configuration.

\begin{figure}
    \centering
    \includegraphics[width=\columnwidth]{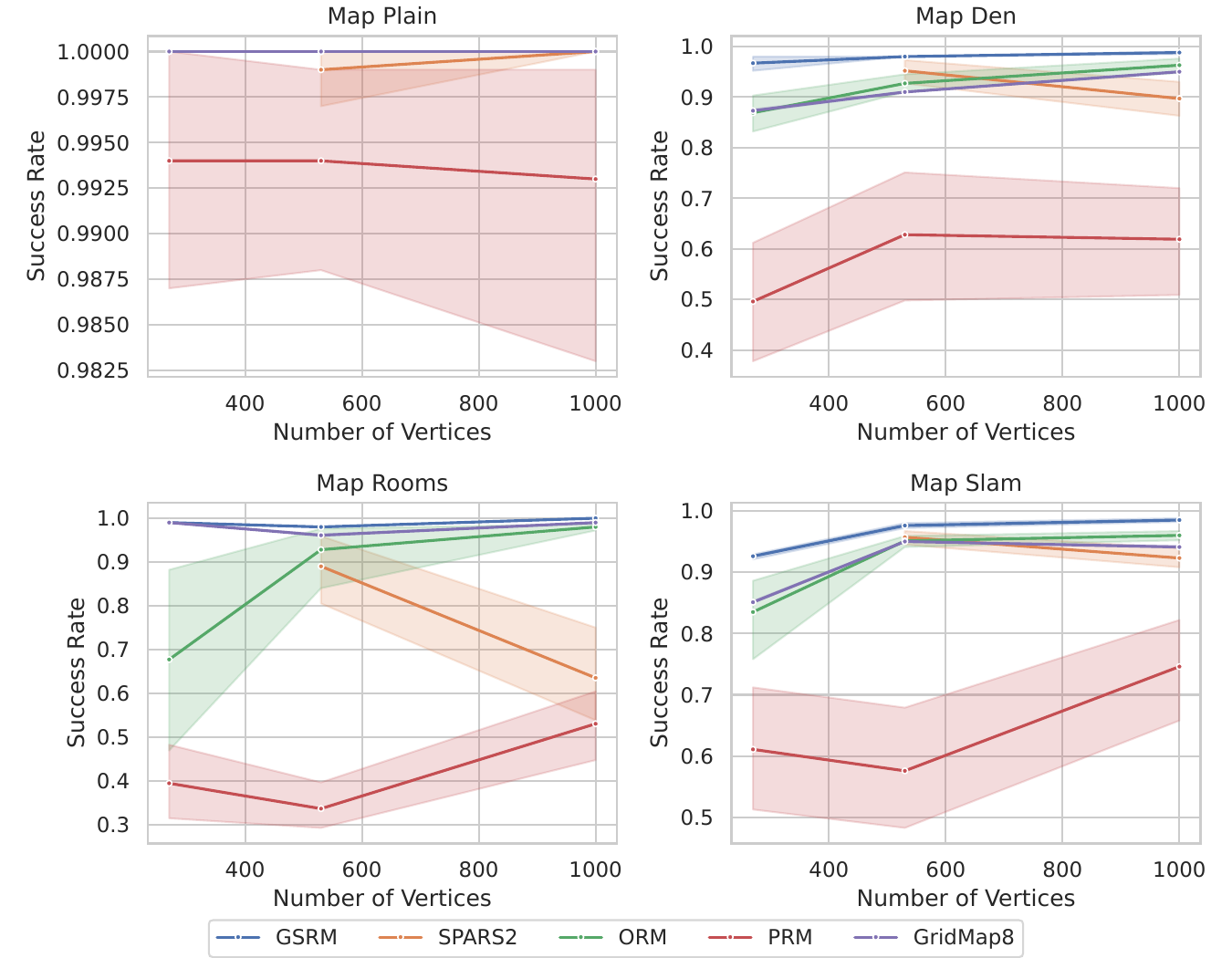}
    \caption{Success rate of the different algorithms for the path planning. A success rate of $1.0$ means that all $100$ paths were successful. It is visible that the \ac{GSRM} algorithm has the highest success rate in all maps.}
    \label{fig:gsrm-success-rate}
\end{figure}

\cref{fig:gsrm-success-rate} shows the success rate of the different algorithms for the path planning.
The \ac{GSRM} algorithm has the highest success rate in all maps.
This can be the fact that other roadmaps have a high likelihood of being disconnected on more structured maps.
It is visible that for most roadmaps, the success rate increases with the number of vertices.
The benefit of the \ac{GSRM} algorithm is that it can generate roadmaps with a high success rate even with a low number of vertices.

Also for the \ac{GSRM} roadmap, the success rate is not  always $1.0$. 
For example on the map \emph{Slam} with the lowest number of vertices.
This can be explained by the roadmap not fully expanding to the small areas of the map that are also visible in \autoref{fig:gsrm-example-last}.
But since we are sampling these for comparability with a fixed number of vertices, these vertices are needed to cover the wider areas of the map and can't penetrate into smaller areas.

\subsection{Query Efficiency (Online Runtime)}
\label{sec:gsrm-evaluation-visited-vertices}
To evaluate the query efficiency of the roadmap we use the metric of visited vertices during the A* path planning algorithm when planning a path between two random points in $\Cfree$. 
This metric serves as a comparable measurement of efficiency during the query phase, as the path planning process is quicker when it expands fewer vertices.

\begin{figure}
    \centering
    \includegraphics[width=\columnwidth]{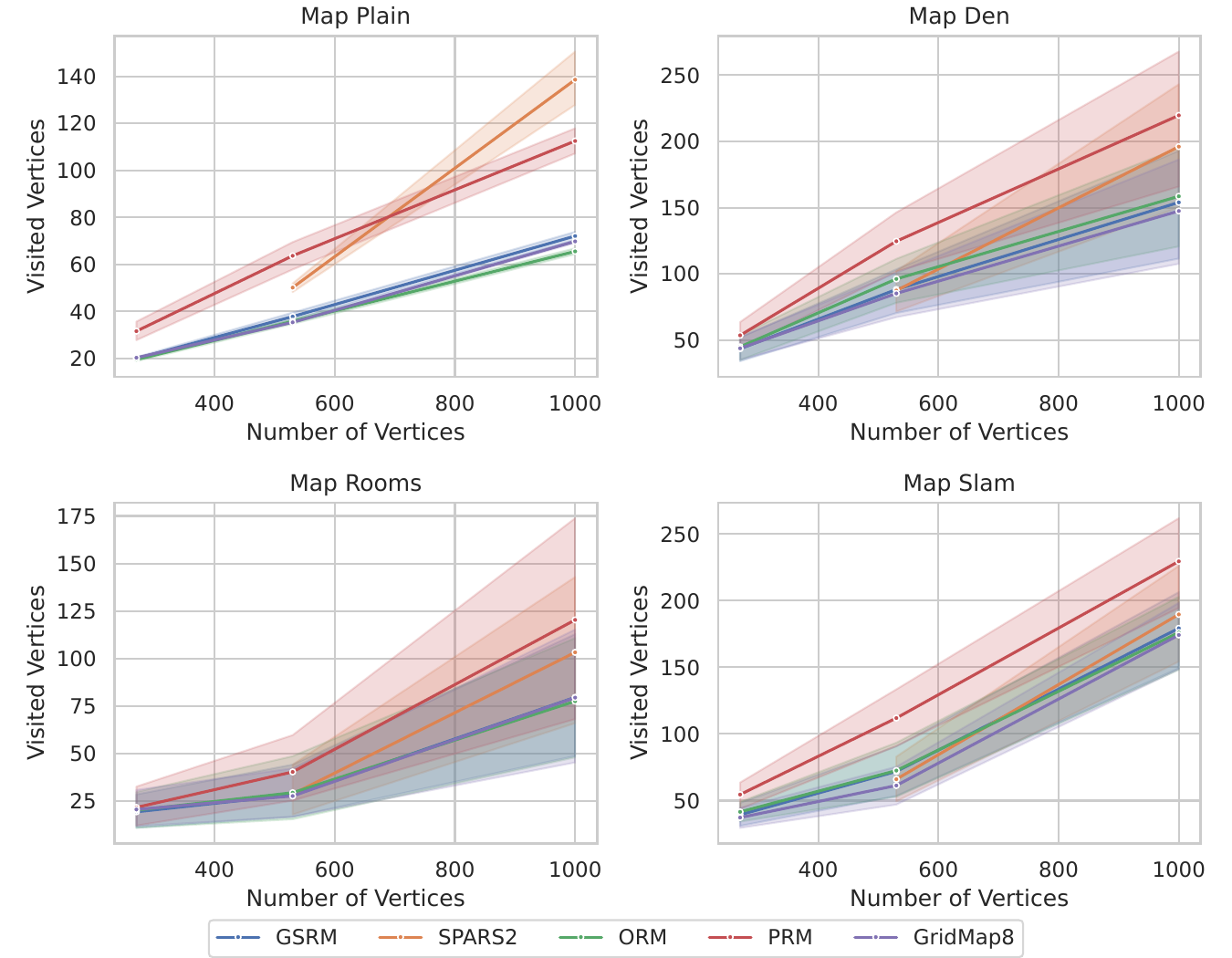}
    \caption{Average number of visited vertices in A* path planning for different roadmap types. The shaded areas show the standard deviation. \ac{GSRM} has a comparable number or smallest number of visited vertices.}
    \label{fig:gsrm-visited-vertices}
\end{figure}

In \cref{fig:gsrm-visited-vertices}, we show the average number of visited vertices in A* path planning for different roadmap types on different maps.
In comparison to the \ac{PRM} algorithm, the \ac{GSRM} algorithm has a similar number of visited vertices to the lowest number of visited vertices of the other algorithms.
It is visible that there is a positive correlation between the number of vertices and the number of visited vertices.
But the \ac{GSRM} algorithm is the lowest for all maps and across different numbers of vertices.


\subsection{Roadmap Construction (Offline Runtime)}
The runtime is an important factor for the applicability of the roadmap generation algorithm.
We measure the runtime on an Intel Core i7-11850H CPU with 32 GB of RAM.
The \ac{GSRM} algorithm is expensive to compute, but its runtime never exceeds 10 seconds on any of the maps up to 2000 vertices.

\begin{figure}
    \centering
    \includegraphics[width=\columnwidth]{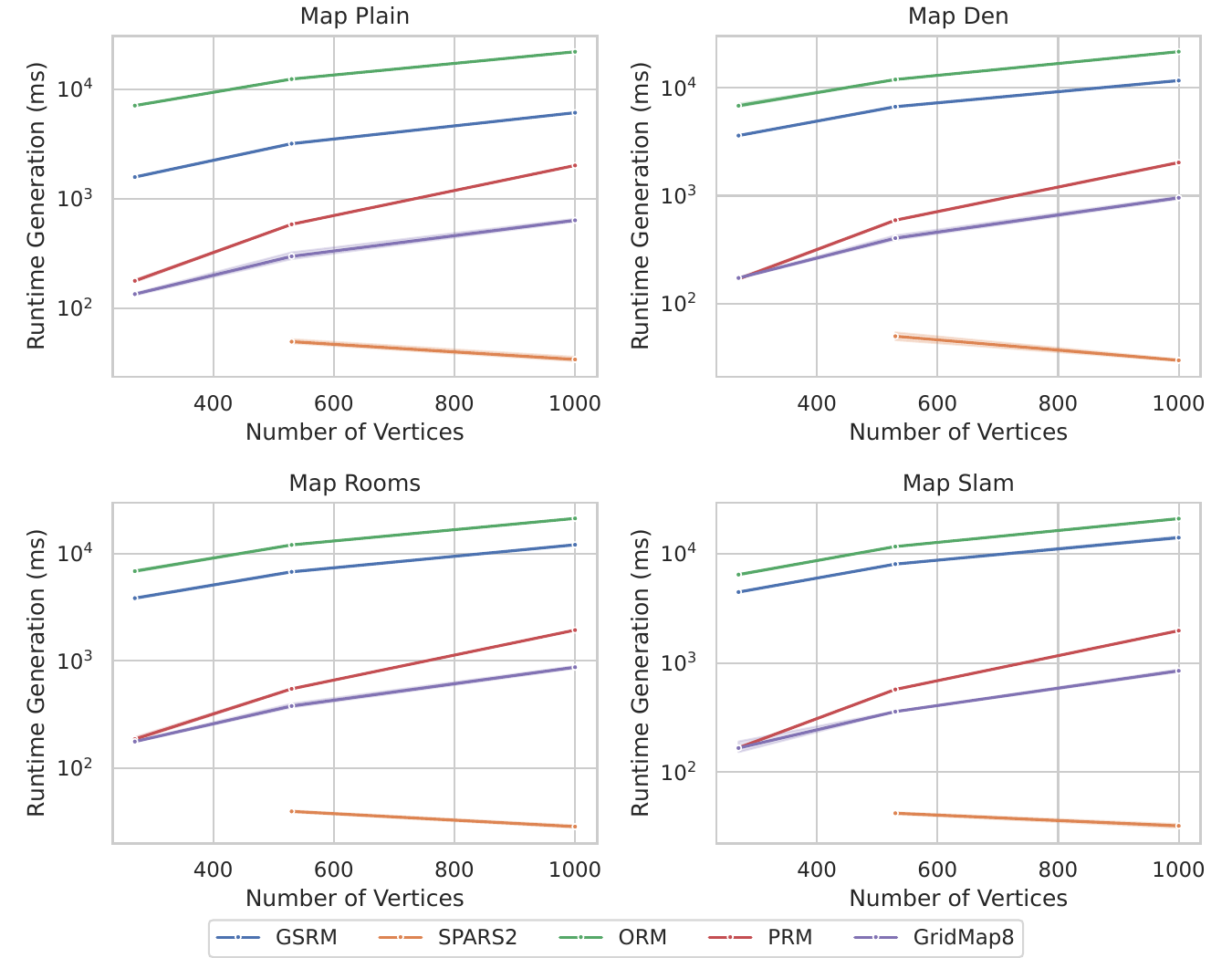}
    \caption{Runtime of the different algorithms for the roadmap generation. The shaded areas show the standard deviation. \ac{GSRM} has the second-slowest runtime.}
    \label{fig:gsrm-runtime}
\end{figure}

\cref{fig:gsrm-runtime} shows the runtime of the different algorithms for the roadmap generation.
Especially, the \ac{SPARS}2 algorithm has a very small runtime.
The only algorithm that is slower than \ac{GSRM} is the \ac{ORM} algorithm.


\section{Conclusion}
\label{sec:gsrm-remarks}
We present GSRM, a new algorithm for roadmap generation for path planning of mobile robots.
Algorithmically, we combine reaction-diffusion systems and triangulation.
Empirically, GSRM exhibits a good tradeoff between path-length, query-efficiency, and computational effort compared to the state-of-the-art.
In future work, we plan to extend the algorithm to 3D and to evaluate it in multi-agent scenarios with distance constraints.


\bibliographystyle{IEEEtran}
\bibliography{z}

\end{document}